\documentclass{article}

\usepackage{arxiv}

\usepackage[utf8]{inputenc} 
\usepackage[T1]{fontenc}    
\usepackage{hyperref}       
\usepackage{url}            
\usepackage{booktabs}       
\usepackage{amsfonts}       
\usepackage{amsmath}
\usepackage{mathtools}
\usepackage{nicefrac}       
\usepackage{microtype}      
\usepackage{graphicx}
\usepackage{caption}

\newcommand{\R}{\mathbb{R}}

\newcommand{\norm}[1]{\left\lVert#1\right\rVert}
\DeclareMathOperator*{\argmin}{arg\,min}
\DeclareMathOperator*{\argmax}{arg\,max}
\DeclarePairedDelimiter{\ceil}{\lceil}{\rceil}

\title{A Survey of Embedding Space Alignment Methods for Language and Knowledge Graphs}

\author{ \href{}{\hspace{1mm}Alexander Kalinowski} \\
	College of Computing and Informatics\\
	Drexel University\\
	Philadelphia, PA 19104 \\
	\texttt{ajk437@drexel.edu} \\
	\And
	\href{}{\hspace{1mm}Yuan An} \\
	College of Computing and Informatics\\
	Drexel University\\
	Philadelphia, PA 19104 \\
	\texttt{ya45@drexel.edu} \\
}

\renewcommand{\headeright}{}
\renewcommand{\undertitle}{}

\hypersetup{
pdftitle={A Survey of Embedding Space Alignment Methods for Language and Knowledge Graphs},
pdfsubject={q-bio.NC, q-bio.QM},
pdfauthor={Alexander Kalinowski, Yuan An},
pdfkeywords={First keyword, Second keyword, More},
}

\begin{document}
\maketitle

\begin{abstract}

Neural embedding approaches have become a staple in the fields of computer vision, natural language processing, and more recently, graph analytics.
Given the pervasive nature of these algorithms, the natural question becomes how to exploit the embedding spaces to map, or align, embeddings of different data sources.
To this end, we survey the current research landscape on word, sentence and knowledge graph embedding algorithms.
We provide a classification of the relevant alignment techniques and discuss benchmark datasets used in this field of research.
By gathering these diverse approaches into a singular survey, we hope to further motivate research into alignment of embedding spaces of varied data types and sources. 
\end{abstract}


\section{Introduction}

The purpose of this survey is to explore the core techniques and categorizations of methods for aligning low-dimensional embedding spaces.
Projecting sparse, high-dimensional data sets into compact, lower-dimensional spaces allows not only for a significant reduction in storage space, but also builds dense representations with many applications.
These embedding spaces have become a staple in representation learning ever since their heralded application to natural language in a technique called word2vec, and have replaced traditional machine learning features as easy-to-build, high-quality representations of the source objects. 
There has been a wealth of study around techniques for embedding objects, such as images, natural language and knowledge graphs, and many research agendas focused on mapping one embedding space to another, either for the purpose of aligning and unifying to a common space, applications to joint downstream tasks or ease of transfer learning. 
In order to fully leverage these dense representations and translate them across domains and problem spaces, techniques for establishing alignments between them must be developed and understood.
To this extent, we believe this avenue of research will continue to blossom, motivating us to present this survey of current methods for alignment of representations of both text and knowledge graphs, classifying them into a taxonomy based on their mathematical approaches and requisite levels of parallel data sources.

Before diving into a survey of alignment techniques, let us first consider their motivation. 
In today's age, the availability of data is pervasive. 
In machine learning, however, methods for learning from this data are skewed to those of supervised learning where labeled data instances are required to optimize models and generalize to new applications. 
Labeled data is a bottleneck for the majority of machine learning projects in industry, and while methods for eliminating this bottleneck have been proposed with degrees of success~\cite{snorkel}, the luxury of labeled data is not always available without prohibitive cost.

At the same time, methods for learning a lower-dimensional representation of data in an unsupervised way have proven useful as inputs to machine learning algorithms. 
These representation learning algorithms, or embeddings, have become a de-facto approach for generating dense and compact feature sets, eliminating the need for tedious human engineering of features at the onset of every new task. 
The success of these techniques is not only related to their proven accuracy in downstream tasks, but their ability to train without supervision, thereby allowing them to scale to massive datasets. 

Given the lack of available supervised data and the prevalence of strong unsupervised methods for embedding large datasets, we consider the problem of matching between embeddings of one set to another, henceforth referred to as embedding alignment. 
We believe that alignment methods provide a convenient method for deriving correspondences between pairs of embedding spaces, thereby providing a method for bootstrapping fuzzy labels between the source and target spaces. 
This problem has applications in, but not limited to, the following areas.

\subsection{Language Translation}
\label{sec:language-translation}
Given a word token in a source language, how can we find a translation of that token in a target language? 
One such way would be to define pairs of tokens in both the source and target language, creating a labeled dataset of translations. 
Building such a dataset would require numerous human-hours and thus may not scale well to, or even be feasible in, languages with limited lexical resources. 
To avoid building these hand-labeled datasets, the task of Bilingual Lexical Induction (BLI) aims to learn mappings from a source to target language in an unsupervised or semi-supervised manner~\cite{bli-survey-penn}. 
A critical task in machine translation (MT) systems, BLI has been influenced heavily by embedding techniques, beginning with linear maps from one embedding space to another~\cite{exploit-sim}. 
By finding structural similarities between two monolingual embedding spaces, such linear maps could generalize to new vocabulary tokens and aid in automated translation.  
Mappings between languages can also be utilized for transfer learning, where a model using embeddings as input features in one language can be re-adapted to another by simply aligning the feature spaces and re-applying the model.
The field of cross-lingual word embedding alignment has rapidly grown, both in the number of evaluation benchmarks and strategies, extending past basic linear maps into the realm of deep learning models~\cite{survey-cross-lingual}.
In this survey, we consider word-to-word, word-to-sentence and sentence-to-sentence alignment tasks as solutions to language translation problems.

\subsection{Knowledge Integration}
\label{sec:knowledge-integration}

An ontology is a formal specification of the types of objects and relationships between those objects in a given domain.
Ontologies are typically developed by ontology engineers with the goal of providing a controlled vocabulary that can be used and reused to provide exact, specific definitions that may be leveraged by humans and machines alike. 
Given that ontologies are domain specific, we may encounter cases where we wish to merge ontologies across two connected domains, or cases where two ontologists have independently developed ontologies for the same domain. 
In such cases, techniques and technologies for highlighting similarities and resolving conflicts are required. 
A goal of ontology integration is to develop a function that takes two ontologies as inputs and output as one merged and aligned ontology, matching like entities and relations to reduce duplicity and resolve entity ambiguity~\cite{ontology-mapping}. 
Recently, ontologies have been popularized as `knowledge graphs' through a clever re-branding by Google~\cite{knowledge-vault}.
Knowledge graphs explicitly frame ontologies using graph data types, allowing for advanced data representation algorithms, such as graph embeddings, to be applied for a variety of tasks, including knowledge base completion~\cite{nickel-survey}.
Framing ontology integration as a problem of aligning embeddings of distinct knowledge graphs, researchers have developed techniques for aligning entities in separate graphs for the purpose of forming a joint graph, many of which are surveyed in this paper~\cite{kg-align-survey}.
In this survey, we consider graph-to-graph alignment as a solution to knowledge integration tasks.

\subsection{Text Understanding and Reasoning}
\label{sec:text-reasoning}

For consumer-facing products like online chat support and information retrieval systems, it is important to have an intuitive interface where a non-technical audience can pose a query and be served information relevant to their needs.
Chatbots, question answering (QA) and retrieval systems alike need to be able to interpret user input into a series of intents, understanding the semantic roles of those intents, as well as parse out syntactic language clues such as negation.
We call this broad umbrella of applications text understanding and reasoning.
Many information reasoning mechanisms are built into knowledge graphs, allowing for the traversals through edges to arrive at facts and inferences.
However, access to those systems typically requires technical expertise.
In order to leverage the power of knowledge graph semantic reasoners, techniques need to be adapted for understanding, including, but not limited to, slot filling, semantic role labeling and sentence analogies.
Harmonizing free text input and understanding with information in a knowledge graph can be seen as an alignment between these two resources, helping to improve the usability and accuracy of these systems.
In this survey, we consider word-to-graph and sentence-to-graph alignments as potential solutions to these tasks. 

\subsection{Information Extraction}
\label{sec:info-extraction}

The ability to turn unstructured text data into structured, machine-operable data is a key motivation behind natural language processing tasks such as part-of-speech tagging, named-entity recognition and relation extraction, all of which can be seen as belonging to the task of information extraction. 
The structured information extracted can be used as inputs to many downstream tasks such as question answering, fact verification and human-computer interactions through chatbots. 
Here, we focus on the task of relation extraction: the ability to identity two entities and the relation between them from raw text data. 
The issue of supervised dataset construction is especially detrimental to research in this area; not only are labeled instances hard to collect as human labelers are known to have low precision for the task~\cite{extreme, extreme-mr}, especially in domains such as biomedicine where subject matter experts are required, but the speed at which new entities and relations may be discussed outpaces the development of such datasets, making supervised models stale and unable to generalize quickly.
To combat the  problem of data collection, the technique of distant supervision was introduced~\cite{Mintz_distantsupervision} allowing for entity and relation triples from an ontology or knowledge graph to be used for quick label generation over raw text inputs. 
In this paradigm, when two entities related by a relation in the knowledge graph appear in a sentence, that sentence is labeled as being representative of that relation. 
This `distant supervision' assumption and its relaxations allow for quick bootstrapping of labeled datasets, but it is well known that they also introduce a great deal of noise, as is the case when a sentence mentioning to entities expresses a novel relation, causing a false label, and are equally susceptible to missing labels when the surface forms don't match or are ambiguous~\cite{dss2019, noise-reduction}.
While distant supervision techniques still depend largely on linking the ontology and text corpus via surface forms (i.e.\ matching on likely string spans or candidate mentions), we anticipate a growing field of alignment between ontology and language embeddings given the increases in alignment techniques used in the two distinct data domains, a main motivating factor for the undertaking of this survey.
The task of information extraction serves as our main motivation for undertaking this survey.
In this light, we focus our efforts on describing techniques or gaps in current research related to graph-to-sentence alignments as this closely mirrors the task of distant supervision.

\subsection{Outline of Survey}
With these motivations in mind, we proceed by first presenting several of the basic methods for generating embedding spaces for language and knowledge graph data in Section~\ref{sec:embeddings}. 
We then move on to describe methods for aligning these spaces in Section~\ref{sec:alignment}, presenting six situations in which alignments are useful and discussing existing research in these areas, where applicable. 
We use Section~\ref{sec:alignment-learning} to provide a categorization of approaches to embedding alignment learning. 
Section~\ref{sec:benchmarks} discusses benchmark datasets used in this area of research. 
We conclude in Section~\ref{sec:conclusion} with a summary and areas of future research.

\section{Embedding Models} 
\label{sec:embeddings}

In this section, we outline the basic methods for embedding language and knowledge data into low-dimensional vector spaces.

\subsection{Word Embedding Models}

Words can be viewed as an atomic unit of natural language. 
Taking this viewpoint, creating features for machine learning models that involve language can be time consuming.
These features typically can include one-hot representations of words or counts of words involved in a sentence or document, both of which suffer from high-dimensionality and sparsity.
Finding dense, lower-dimensional representations of words to replace traditional features is the focus of work on word embeddings.
The first such modern approach was word2vec, an approach that leverages a shallow neural network to generate hidden state representations. 
By training such a network, co-occurrences between words are learned to project like-words into the same areas of the low-dimensional space, reflecting their syntactic and semantic properties. 
Two similar formulations, the continuous bag-of-words model (CBOW) and Skip-gram models were proposed by~\cite{word2vec}; here, we focus on the Skip-gram model.
We first define the probability of a word $w_i$ given another word $w_j$ as 
\begin{center}
$$p(w_i|w_j) = \frac{\exp(u_{w_i}^\top v_{w_j})}{\sum_{l=1}^V \exp(u_{w_l}^\top v_{w_j})}$$
\end{center}
where $u_w$ is the trainable vector of input probabilities and $v_w$ the trainable vector of output probabilities for a given word $w$, and $V$ is the entire vocabulary of the given language domain. 
Given that the size of $V$ is typically very large, these probabilities are estimated by leveraging negative sampling where noise words are used to generate large contrast against potential high probability guesses, eliminating the need to estimate these probabilities over the entire vocabulary for each word.
With a sequence of words $w_1, w_2, \ldots , w_n$, the goal of the Skip-gram model is to maximize
\begin{center}
$$\frac{1}{T} \sum_{t=1}^T \sum_{j=-k}^{k} \log p(w_{t+j}|w_t)$$
\end{center} 
where k specifies the window size, i.e.~how far to the left and right of the centered word $w_t$ we look when calculating the probability.

Advances and improvements over this model are plentiful.
In~\cite{glove} the authors develop the GloVe model, making use of a global co-occurrence matrix to address information lost by focusing on small windows during the training of word2vec models.
To make the training robust to spelling errors and easier to apply to unseen words at training time, the authors of~\cite{fasttext} build FastText, using sub-word embeddings made up of units of characters.
Addressing the problem of polysemy, bi-directional neural network models are utilized to capture further context about word usage.
Using the Transformer module of~\cite{all-you-need}, word embeddings have matured from \textit{static} representations of those in word2vec, GloVe and FastText, to \textit{contextual} representations used in BERT~\cite{bert}, ELMo~\cite{elmo} and GPT-2~\cite{gpt-2}.
These models are massive in neural architecture, capturing long distance dependencies between words and interactions of words in multiple contexts.
Because of their size, they have the requirement of very large training sets.
Due to data and architecture size, these models are typically pre-trained, where initial weights are learned and shared, then re-trained, or fine-tuned, for target tasks or datasets.
The availability of these pre-trained models though several APIs has made their use the status quo for natural language processing tasks.

\subsection{Sentence Embedding Models}\label{sec:sentence-embedding}

Building on the successes of word embedding models, a logical next step is to use words as atomic units that are composed into sentences.
In this shift, moving from a discrete world where words typically represent a handful of semantic units to a continuous representation in a sentence or document, where words can be combined in infinitely many ways, represents a significant challenge.
Beginning with sub-word embeddings trained using FastText, an efficient sentence classification model is established in~\cite{fasttext} by representing a sentence as the average of its component word representations.
Taking the average or sum of static vectors is a common approach to move from word representations to sentence representations, as discussed in~\cite{sentence-analogies}, and can often beat more advanced models while retaining a level of simplicity.
Even though these representations provide successful baselines, they throw out an important element of data in moving from words to sentences: word ordering.
To address this, ~\cite{discretecosine} propose utilizing a discrete cosine transformation.
By stacking the individual word vectors $w_1, \dots , w_N$ into a matrix, the discrete cosine transformation can be applied column-wise: for a given column $c_1, \ldots , c_N$, a sequence of coefficients can be calculated as 
\begin{center}
$$coef[0] = \sqrt{\frac{1}{N}}\sum_{n=0}^N c_n$$
\end{center}
and
\begin{center}
$$coef[k] = \sqrt{\frac{2}{N}}\sum_{n=0}^N c_n \cos \frac{\pi}{N} (n + \frac{1}{2})k$$
\end{center}
The choice of $k$ typically ranges from 0 to 6, where a $k$ of zero is essentially the same as vector averaging, while higher orders of $k$ account for greater impacts of word sequencing. 

Alternative approaches abandon static word vectors and focus on the sequence of words in the sentence as the starting point.
To accommodate sequential data, recursive neural networks (RNNs) dominated the field for a period of time.
Recurrent neural network architectures provide an added benefit in that they can theoretically process sequences of variable length up (in practice, this is up to some max length where other sequences are padded with a special token), allowing them to train on corpora with long and short sentences.
Another key advantage of RNNs is their ability to share parameters over time where signal from a prior word carries forward to the next word, and so on.
This benefit of carrying information forward through the network has a downside of making them hard to train, as gradients need to be propagated backward through time; this has caused them to fall out of favor.
One of the first such RNNs trained for sentence encoding was the Skip-Thought model~\cite{skipthought}.
Rather than use word-context windows, the Skip-Thought model generates an encoding for a center sentence and uses that encoding to predict $k$ sentences to the left and right, where $k$ is again the window size as in the Skip-gram model.
To accomplish this, the model leverages an encoder-decoder architecture where each encoder step takes the sequence of words in the sentence and represents them as a hidden state, which is then encoded through the RNN.
Decoding then takes place in two steps, one for predicting the next sentence and one for the prior sentence, each of which generates a hidden state through time that can be used to calculate the probabilities of each word in the sequence, with the following objective function
\begin{center}
$$\sum_t \log P(w_{i+1}^t| w_{i+1}^{<t}, h_i) + \sum_t \log P(w_{i-1}^t| w_{i-1}^{<t}, h_i)$$
\end{center}

An extension of Skip-Thought is the Quick-Thought model~\cite{quickthought}.
The authors note that the objective function used in Quick-Thought is focused on re-creating the surface forms of each sentence given its dependence on the individual words represented.
Specifically, the authors claim ``there are numerous ways of expressing an idea in the form of a sentence.
The ideal semantic representation is insensitive to the form in which meaning is expressed''~\cite{quickthought}.
The objective function of Quick-Thought is thus changed to focus only on sentence representations, using a discriminative function to predict a correct center sentence given a window of context sentences.
The authors of~\cite{infersent} continue on the quest to capture sentence semantics with the InferSent model.
InferSent uses a supervised learning paradigm where sentences in the training set are fed into a three-way classifier, predicting the degree of their similarity (similar, not similar, neutral).
Coupled with a bi-directional LSTM model, the InferSent model can be pre-trained on natural language inference (NLI) tasks such as sentence semantic similarity and later used for inference or fine-tuning on other tasks.
 
Similar to the InferSent model, Sentence-BERT uses supervised sentence pairs to learn a similarity function~\cite{sentencebert}.
The input sentence embeddings used for the three-way similarity classifier are generated from a pre-trained BERT model.
Contextual models such as BERT provide an encoding of each positional word in an input sentence as their output, thus it is necessary to aggregate these contextualized representations into a single static sentence representation.
As with word embedding models, this aggregation can be a sum or average of the representations at a particular layer of the language model, typically the top layer or final layer.
An alternative option is to provide the model a special classification token ``[CLS]'' that has been pre-trained to compress the contextual representations into one layer, which can then be fed to a non-linear unit.
Sentence-BERT also adapts the classification task to one of regression where cosine similarity scores are used to score the degree of similarity between sentences.
This approach is useful for particular applications, such as semantic search.

Continuing on the path of larger, deeper architectures powered by more data, ~\cite{laser} train a Bi-directional LSTM model on a massive scale, multilingual corpus to generate sentence embeddings.
Using parallel sentences across 93 input languages, the authors were able to focus on mapping semantically similar sentences to close areas of the embedding space, allowing the model to focus more on meaning and less on syntactic features.
Each layer of the LASER model is 512 dimensional, with an output concatenation of both the forward and backward representation generating a final sentence representation of dimension 1024.
The model outperforms BERT-like architectures for a variety of tasks including cross-lingual natural language inference, a task focused on detecting sentence similarities. 
 
\subsection{Knowledge Graph Embedding Models}

Building on the success of embedding-based methods in natural language processing, these techniques have spilled into the domain of knowledge graphs.
Their main motivating uses are in the task of statistical representation learning where the larger graph is compressed into a low-dimensional representation that can be used by reasoning systems, and knowledge base completion (KBC), where embeddings of existing facts can be utilized to predict new relationships between entities in the graph.
Approaches in this area can be classified into three main categories: translation-based models, semantic-matching models and graph-structure models. 
Our aim is to introduce models leveraged in the alignment literature; for more comprehensive introductions see~\cite{nickel-survey} and~\cite{ke-survey}.

\subsubsection{Translation-based Methods}

Let $G = (E, R)$ be a knowledge graph consisting of a set of entities $E$ and relations $R$, each element of which may have an entity or relation type. 
From this graph, we can construct the set of known facts, represented as triples $\langle h, r, t \rangle$ with $h,t \in E$ and $r \in R$.
The intuition behind translation-based models is we wish to have low-dimensional, dense representations of $h, r, t$ such that $h + r \approx t$.  
Model choices then depend on which space or spaces the entities and relations are embedded in as well as the scoring function used to help the model learn to differentiate between true triples from the graph and noise triples that do not reflect real-world facts.
TransE~\cite{transe} is the simplest of these models.
It embeds both the entities and relations in the same low-dimensional vector space and uses a simple distance function defined by
\begin{center}
$$f_r(h,t) = -\norm{h + r - t}_{1/2}$$
\end{center} 

While this model is simple, it struggles to properly encode one to many triples, where a single relation may hold between a head entity and several tail entities.
Resolutions to this are addressed in TransH~\cite{transh}, where each relation is assigned its own hyperplane.
Similarly, TransR~\cite{transr} lets each relation have its own distinct embedding space ad thereby greatly expands the parameter space of the model and increases the capability of learning relation-specific translations.
An entire family of translation-based models exists, each adding various complexities and constraints to the embedding spaces with novel loss functions used to best recover the relations described in the original graph.

\subsubsection{Semantic-matching Models}\label{sec:semantic-matching}

While the translational assumption $h + r \approx t$ gives good geometric intuition as to the types of relations learned during model training, it is prohibitive for a wide class of relations, including those with anti-symmetric or complex properties.
Semantic-matching models deviate away from the distance-based assumption and focus on using similarity-based scoring functions to recover facts from the low-dimensional representations of entities and relations.
Rather than relying on norms and translations, these methods leverage dot-product like scoring functions to measure angles between low-dimensional representations, sometimes referred to as `semantic energy' functions.
The simplest of such models is RESCAL~\cite{rescal}, which relies on a tensor representation of the underlying knowledge graph $X$, where each entry of the tensor $X_{ijk} = 1$ if the fact is represented in the knowledge graph, else-wise zero.
This tensor can then be factorized into latent components,
\begin{center}
$X_{k} \approx A R_k A^\top$ for $k = 1, \ldots , r$
\end{center} 
where $R_k$ is a matrix of dimension $r \times r$ representing interactions between each corresponding component and $A$ contains the $r$ dimensional representations of the entities. 
Thus for each $(h,t)$ pair, we can compute the likelihood they participate in the $k$-th relation as
\begin{center}
$$f_k(h, t) = h^\top R_k t$$
\end{center}
Contrasted with translation-based models, RESCAL takes advantage of vector products while capturing interactions between elements of each entity and all relations. 
In a simplification, DistMult~\cite{distmult} requires each $R_k$ to be diagonal, reducing the parameters of the model while sacrificing some of its representational capacity.
This reduction in capacity is especially felt when modeling anti-symmetric relations as interactions in these diagonal matrices have no notion of directionality.
To circumvent this issue, the ComplEx~\cite{complex} model allows for the low-dimensional representations to live in the complex space $\mathbb{C}$.
The scoring function used by the ComplEx model is defined as
\begin{center}
$f_k(h, t) = \Re(\langle h, w_k, t \rangle)$ where $w_k \in \mathbb{C}^r$.
\end{center}
By allowing the representations to be complex-valued, the model can handle the asymmetries of many relations present in knowledge graphs, yet score the likelihoods of facts existing using only the real-valued vectors.
The work of~\cite{conve} takes this one step further, defining the ConvE model where entities interact through the convolution operator. 
This introduces additional non-linearities through which the model can increase the capacity for learning complicated relational structures. 

\subsubsection{Graph-structure Models} 

Given that the entities and relations between them are modeled as a graph with vertices and edges, we can take advantage of the underlying graph structure to aide in creating low-dimensional representations.
Graph representation learning has been a trending topic over the past few years with many advances in creating representations of graphs that can be used in machine learning models.
We refer the reader to~\cite{rlg-survey} for a complete introduction.

For knowledge graph embedding, path traversal techniques have been applied to learn additional facts about the relations between multiple entities in a graph, rather than just the one-hop paths learned in translation-based and semantic-matching models.
By taking a walk on the graph we can learn more about the neighborhood structures of each entity, using that information to learn better dense representations.
Using paths on the graph, the PTransE approach extends the traditional TransE method to capture structural information~\cite{ptranse}.
Given two entities $h$ and $t$ and a path $p = r_1 \rightarrow r_2 \rightarrow \ldots \rightarrow r_k$, where each $r_i$ is a relational embedding, the authors of PTransE propose three ways of aggregating all relation vectors involved to a path vector.
These include addition: $p = r_1 + \ldots + r_k$, multiplication: $p = r_1 \cdot \ldots \cdot r_k$ and application of a RNN: $c_i = f(W[c_{i-1}\colon r_i])$ where $f$ is a non-linearity and $[:]$ represents vector concatenation.

In a similar approach,~\cite{traversing} create entirely new triples from paths in the knowledge graph.
If $h$ and $t$ are connected by the path $p = r_1 \rightarrow \ldots \rightarrow r_k$, they add a new triple $\langle h, p, t \rangle$ to the set of known triples used to train the knowledge graph. 
The triples can now be recovered using the translation-based loss of TransE
\begin{center}
$f_k(h,t) = -\norm{h + (r_1 + \ldots + r_k) - t}_{1/2}$
\end{center}
or used in the RESCAL context through multiplication of the relevant slices of the factorized matrix $R_k$
\begin{center}
$f_k(h,t) = h^\top (M_1 \cdot \ldots \cdot M_k) t$
\end{center}
where each $M_i \in R_k$.

More recently, attention has turned to using graph convolutional networks~\cite{kipf-gcn} (GCNs) for knowledge graph embeddings.
These techniques are called convolutional as they use neighborhood features of each node, similar to how convolutional operators look at borders of each pixel in computer vision models.
By representing the knowledge graph $G$ by its adjacency matrix $A$ and let $X$ be a matrix of representations (features) of the entities in the graph.
The convolutions, defined layer by layer, can be represented as
\begin{center}
$H^{(l+1)} = \sigma (\tilde{D}^{-1/2} \tilde{A} \tilde{D}^{-1/2} H^{(l)} W^{(l)})$
\end{center}
 for the $l+1$th layer, where $\tilde{A} = A +I_N$ and $I_N$ is the identity matrix, $\tilde{D_{ii}} = \Sigma_j \tilde{A_{ij}}$and $W^{(l)}$ is the weight matrix for each layer.
Here, $H^{(0)} = X$, meaning the process begins by considering individual nodes and expands to represent their neighborhoods up to the number of layers in the network.

Translation-based methods, semantic-matching models and graph-structure models have all been used to embed individual knowledge graphs as well as aide in entity alignment between embedded graphs, as described in Section~\ref{sec:kg-to-kg}.

\section{Alignment Approaches}
\label{sec:alignment}

Abstractly, learning a mapping function between two vector spaces is a well studied problem. Let $\mathcal{D}_1$ and $\mathcal{D}_2$ be two datasets, originating from either similar (as is the case for two language corpora from different languages) or different (as is the case for a set of images and a language corpus) domains.  
Let the functions $f_1\colon \mathcal{D}_1 \rightarrow \R^n$ and $f_2\colon \mathcal{D}_2 \rightarrow \R^m$ represent two mappings from the original datasets to their respective real-valued embedding spaces. 
Typically, $n$ and $m$ are of much lower dimension than the original cardinalities of $\mathcal{D}_1$ and $\mathcal{D}_2$, and therefore $f_1$ and $f_2$ can be thought of as techniques to compress the original datasets whilst maintaining their defining geometric characteristics, including a notion of `semantic similarity'. 
These similarities are measured in the lower-dimensional vector spaces through techniques such as, but not limited to, Euclidean distance or cosine similarity.

Let us assume that these `semantic similarities' are preserved by the functions $f_1$ and $f_2$. 
If there exists a correspondence between elements $x \in \mathcal{D}_1$ and $y \in \mathcal{D}_2$, then the problem of aligning their respective embedding spaces seeks to find a map $A\colon \R^n \rightarrow \R^m$ such that $A(f_1(x)) \approx f_2(y)$.

More generally, these methods seek to detect and exploit \textit{invariances} between pairs of low-dimensional embedding spaces. 
The degree to which these invariances can be captured dictates how much training data is required to learn a reliable alignment model. 
In the case where the underlying geometric structures of both embedding spaces are perfectly invariant, up to a rotation of the space, simple maps may be learned in a highly unsupervised way. 
However, on the flip-side of the coin, methods which do not generate well structured embedding spaces may require more training data in order to learn alignments. 
Critically, the problem of learning an alignment map $A$ is also tied to the choice of good embedding functions $f_1$ and $f_2$, and careful coordination between all three choices is required for finding an optimal solution.

\subsection{Research Landscape}\label{sec:landscape}

As mentioned in our motivating section, we believe that embedding alignments are critical to the task of information extraction, particularly in mapping unstructured text to structured knowledge graphs.
However, our hypothesis is that these techniques have yet to be fully explored in the research community, especially in learning alignments between sentence embeddings and graph embeddings.
To understand the landscape of research in alignments, we undertook a search of three research repositories: ArXiv, DBLP and IEEE.
Our search dates ranged from January 1, 2012 (picked to cover a period of a year before the publication of the word2vec paper) through the end of 2019 (to avoid a partial year of 2020 at the time of publication).
For each repository, we conducted a keyword search for `embedding alignment'.
We then further sub-divided the results into four categories as they pertain to embeddings: knowledge graph, sentence, word or not applicable.
These categories are determined by a simple count of keyword matches in the paper's abstract, as outlined in~\ref{Tab:landscape-key}, with ties being assigned to both categories.

\begin{center}
\begin{table}
\captionof{table}{Keyword Labels for Research Classification \label{Tab:landscape-key}}
\centering
\begin{tabular}{| c | c |}
 \hline
 Label & Keywords  \\
 \hline
  Knowledge Graph & node, knowledge graph, network, ontology, knowledge base  \\
  Sentence & sentence, phrase, cross-lingual, multilingual \\
  Word & word, token, cross-lingual, multilingual  \\
 \hline
\end{tabular}
\end{table}
\end{center}

We then measured the trend over time for mentions of word embedding alignments, sentence embedding alignments and knowledge graph embedding alignments. 
The results are shown in the following figure.

\begin{center}
\includegraphics[scale=0.65]{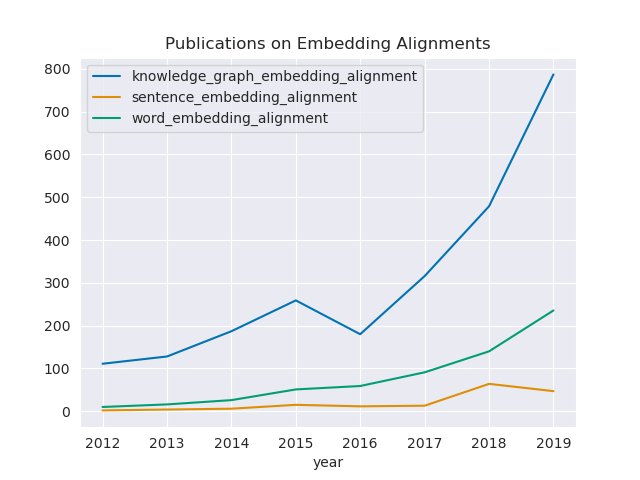}
\end{center}

We note that the inclusion of the `network' keyword artificially inflates the count of publications classified as aligning knowledge graph embeddings.
Many of these papers deal with embeddings of social networks, and while those networks could be considered knowledge graphs they do not fit the definition of a knowledge graph used herein.
The same trend plot with network related papers removed can be seen as follows.

\begin{center}
\includegraphics[scale=0.65]{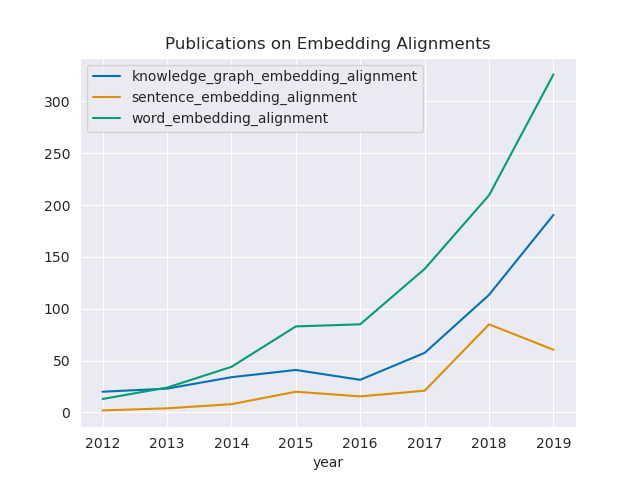}
\end{center}

Per the above trends, we see that there has been a continued rise in the application of both knowledge graph and word embedding technologies.
Mentions of sentence embeddings are dwarfed by the other two categories, with some of that trend explained by limitations in~\ref{sec:sen-sen}.
Given the popularity of embedding techniques in the machine learning community, we believe growth in this research area will continue.
We also hypothesize that alignments between other objects, i.e. tokens to graphs, will become an increasingly important field for knowledge and data integration.
We proceed by enumerating the potential applications of embedding space alignments.

\subsection{Alignment Use Case Enumeration}

Given the focus of this survey on the domains of language and knowledge graphs, we outline six situations in which embedding space alignments could occur. 
In these cases, we assume that the direction of the learned alignment mapping is irrelevant, i.e.~we could easily reverse the source and target spaces and learn an alignment map in the reverse direction.

\subsubsection{Word-to-Word Alignment}

The ultimate goal of a word-to-word alignment model is to be able to input the embedding of a token in a source language and receive as output the embedding of a semantically or syntactically similar token in the target language.
As first noted in~\cite{exploit-sim}, word embedding models trained on distinct languages exhibited similar geometric patterns and behaviors.
This observation led the authors to hypothesize that word embedding spaces could be transformed from one to another through simple linear operations, such as translation and rotation.
The first attempts and models in this area took advantage of large, parallel vocabularies, where pairs of words were used to learn mapping matrices from one space to another.
While learning the transformation matrix may have a closed-form solution and could be directly solved through linear algebraic methods, in practice, the weights are learned through stochastic gradient descent.
We survey the common supervised learning model types in Section~\ref{sec:word-supervised}.
Given the relative success and ease of implementation of these models when parallel data is available, researchers began to ask how limited that parallel set could be.
Restricting to the top 5,000 most common words, restricting the parts of speech, or even relying only on numerals have been popular approaches into reducing the level of supervision needed to learn strong translation models~\cite{biwe}.
Hybrid approaches use a form of semi-supervised learning, beginning from small seed lexicons and iteratively adding words as confidence in their direct translation builds.
We introduce these semi-supervised methods in Section~\ref{sec:word-semi}.
Moving past semi-supervised methods, approaches to learning mappings between embedding spaces in a completely unsupervised way.
These methods rely on the geometric structures of the underlying spaces as a proxy for parallel data, either relying on embedding similarity distributions~\cite{self-learning}, adversarial learning~\cite{non-parallel-conneau} or metric recoveries via optimal transport~\cite{gw-align}.
We cover these methods in Section~\ref{sec:word-unsuper}.

\subsubsection{Sentence-to-Sentence Alignment}\label{sec:sen-sen}
Sentence to sentence alignment often serves as an entry point to machine translation applications.
Given a parallel corpus of sentences in two languages, the goal is to learn a mapping function $f$ that converts a low-dimensional representation of sentence $s_1 \in \L_1$ to a close (in terms of vector space proximity) representation  $t_1 \in \L_2$.
This map can then generalize for future translations such that $f(s_2) \approx t_2, s_2 \in L_1, t_2 \in L_2$. 
This approach is limited due to two factors.
First, the availability of such parallel corpora is limited.
Most research in this area either relies on the Europarl dataset~\cite{europarl} or translations of the Bible.
Neither of these resources represents enough diversity in language to scale up to production-level systems, but they do allow for ideas to be tested experimentally.
The second limitation comes from the composition of semantic units (i.e. individual word tokens) to higher order representations in sentences.
Word order plays a role in the structure of languages, thus simple mapping models have been replaced with those that model the sequences of tokens, such as the seq2seq model~\cite{seqseq}.

\subsubsection{Word-to-Sentence Alignment}
Given the availability of technologies for word-to-word and sentence-to-sentence alignment, there has been little need for additional research in word-to-sentence alignment. 
In the case of a set of sentences being mapped to a finite set of words, this is typically handled as a supervised classification problem where the finite set of words is one-hot encoded to represent a target variable.
We direct the reader to~\cite{fasttext} for efficient approaches to this type of supervised classification problem.

\subsection{Sentence-to-Sentence Alignment}

While the goal of word-to-word alignment is to map tokens for direct translation, these tokens often can express multiple senses and thereby exhibit polysemy.
This creates issues in direct, one-to-one mappings due to the fact that the training data can contain a particular token in the source space with multiple translations in the target space, leading to conflicting information during training and at inference.
Rather than focusing on tokens as the atomic unit, tokens in a given context, either through phrases or complete sentences, carry more information that can be leveraged for better alignment.
The research area of sentence-to-sentence alignment relies on parallel documents, typically found in resources such as translations of the European Parliament proceedings or the Bible.
While alignment of parallel word tokens is a rich research field, there has been less focus on alignment techniques of full sentences; research in this area typically falls under the umbrella of machine translation where more complex sequence-to-sequence neural models are favored.
However, many of the same techniques for aligning word embeddings can be leveraged for aligning sentences provided we can generate representative sentence embeddings.
We cover a handful of sentence embedding methods in~\ref{sec:sentence-embedding} and discuss their alignment in section~\ref{sec:sentence-align}.

\subsection{Knowledge-to-Knowledge Alignment}
\label{sec:kg-to-kg}
Knowledge graphs have seen a great deal of interest and hype in recent years as their applications to artificial intelligence and machine learning have come to be seen as the onset of a `third wave' contributing to semantically grounded and explainable AI.
They also serve as the backbone to the Semantic Web, a set of standards for defining and linking data and meta-data in a machine-readable and human-interpretable way.
While large corporations like Google and LinkedIn have massive knowledge graphs at their disposal, smaller, more tailored graphs exist for specific purposes such as SnoMed for medical clinical terminology and FIBO for financial industry concepts.
Given the specificity of some of these smaller graphs, we may wish to weave several of them together for a particular application, including data exchange protocols and data integration tasks.
We may also want to merge knowledge graphs covering similar, yet independently defined, concepts, or graphs defining the same subject matter across languages.
This task is referred to in the literature as \textit{entity alignment}: the process of identifying nodes in each graph that are referencing the same semantic concept and either forming relations between them or compressing them into a single representation.
Work in this area originated in the task of ontology alignment~\cite{noy-semantic-integration}, which aimed to use heuristics, string matching and natural language processing techniques to map source and target nodes.
As in the other application areas in this survey, the push to deep, representational learning has invaded the space of ontology alignment as well, typically couched under the banner of entity embedding alignment.
The task at hand is to create low-dimensional representations of the source and target knowledge graphs and use only these embeddings to automatically discover alignments.
As in word-to-word alignments, methods range from directly supervised methods where parallel entities between graphs are used to learn mappings, to fully unsupervised methods where inferences are made to align entities based on the structure of their neighborhoods in the graph.
We cover these techniques in Section\ref{sec:graph-to-graph}

\subsubsection{Word-to-Knowledge Alignment}

In the previously explored cases, embeddings of source and target data from similar domains were aligned.
In these cases, strings are mapped to strings and graph entities to other graph entities.
This section deviates from  those proceeding by considering alignments from strings to knowledge graph entities.
For a given token $x \in \mathcal{D}_1$, we wish to identify a corresponding entity $y \in \mathcal{D}_2$, if such an entity exists.
One such way of finding these correspondences is to find a map between the token embedding $f_1(x)$ and the entity embedding $f_2(y)$.
Given that the target domain (a knowledge graph) is constructed to reflect facts about real world entities and the relations between them, we expect to find those same facts and entities referred to in the source space (language), although with much lower precision and exactness in their statements.
While inherent noise present in human language makes learning such an alignment challenging, success in this area can assist with knowledge driven entity extraction and named entity recognition.

\subsubsection{Sentence-to-Knowledge Alignment}

Our main motivation in this line of research is the alignment of sentences to knowledge graphs.
The interest in this problem is two-fold.
Firstly, if we are able to align embeddings of triples $\langle h, r, t \rangle$ from the knowledge graph $G$ to sentence embeddings $s$ in a given corpora, these alignments can be used to detect the expression of relationships $r$ in the sentences, aiding in the task of relation extraction.
Secondly, in the opposite direction, if we can align sentences to triples, we can use this technique to assist in the detection of new triples to be added to the knowledge graph from text data, aiding in the automated expansion of a knowledge graph.
These two problem domains can be viewed as complementary techniques for converting unstructured data in text documents to structured data in a knowledge graph.
Having data in a structured format not only makes it easier for human verification, as in the case of automated fact checking, but also allows for insights into how other machine learning models, such as document classification, are leveraging unstructured data, providing an avenue for explainable AI and model governance.

\section{Alignment Learning Paradigms}
\label{sec:alignment-learning}

In this section, we explore the major techniques used in each of the six categories defined above.
Each section reviews the works from the perspective of classifying them into supervised, semi-supervised and unsupervised frameworks, motivated by our desire to assess the requisite amount of parallel data necessary to learn an alignment.

\subsection{Word-to-Word Alignment Techniques}

We proceed by classifying word-to-word alignment techniques into supervised, semi-supervised and unsupervised methods.

\subsubsection{Supervised Methods}\label{sec:word-supervised}
Supervised learning methods are the most common and most data intensive in machine learning applications. 
To help alleviate the burden on developers of these methods, leveraging unsupervised methods as discussed above helps to ingest large amounts of data and build robust features to jumpstart learning.
In this section we discuss supervised models that use unsupervised features as inputs with the goal of aligning these resources.

\paragraph{Regression Models}
Regression models form the class of solutions first used to address the word-to-word embedding alignment problem. 
Let us begin by defining languages $L_s$ and $L_t$, our source and target language, respectively, and embedding functions $f_1: L_s \rightarrow \mathbb{R}^n$ and $f_2: L_2 \rightarrow \mathbb{R}^m$. 
Given a set of parallel translation tokens $(w_i^s, w_i^t)$ where $w_i^s \in L_s$ and $w_i^t \in L_t$, we wish to learn a transformation matrix $W$ to minimize the following mean-squared loss
\begin{center}
$$\sum_{i=1}^{n}  \norm{Wf_1(w_i^s) - f_2(w_i^t)}^2$$
\end{center}
This method was first proposed by~\cite{exploit-sim} as a means of capturing geometric patterns between embeddings across embedding spaces. 
In the original paper, no additional pre-processing is done on the input word vectors, which were generated using the CBOW word2vec algorithm.
The transformation matrix $W$ can then be applied to a new vector $f_1(w_N)$ to map it into the target space where a cosine similarity search can rank all translation candidates.
Subsequent papers suggested minor tweaks to the regression model having significant impacts on the capacity to learn.
These include the addition of $l_2$ regularization~\cite{hubness} and adding pre-processing steps such as embedding vector unit normalization, further discussed in the following section.

\paragraph{Orthogonal Models}\label{sec:orth-word}
The original regression model utilized a Euclidean distance in learning the transformation matrix, yet relies on cosine similarity to carry out similarity searches in the target space.
This inconsistency was first noted by~\cite{normalized} who in turn modified the regression process to add unit length normalization to the source and target vector spaces and constrain the matrix $W$ to be orthogonal, that is $W^\top W = I$ where $I$ is the identity matrix.
The pre-processing step and orthogonal constraint then line up with the retrieval method, where we are less concerned with distances between vectors and more concerned with the angles between them.
Solutions to this minimization problem are still carried out by stochastic gradient descent where the orthogonality constraint is implemented by solving the SVD problem, typically done by mini-batch fed to the optimizer.

Applications of pre-processing and orthogonal constraints spurred further research into ways to manipulate the source and target embedding spaces to further express their geometric structures.
In~\cite{monolingual-invariance} and~\cite{multistep-linear}, the authors evaluate several pre- and post-processing steps, building towards a framework of applicable methods.
The steps in this framework are enumerated as follows:
\begin{itemize}
\item Normalize the source and target spaces, either using unit norms or mean centering (where each component/feature is forced to have zero mean) as an initial pre-processing step;
\item Feature whitening, requiring each feature to have unit variance and removing their correlations, applied to both source and target space independently;
\item Learning an orthogonal mapping via the regression technique;
\item Re-weight the features to increase their correlations between source and target spaces, only applied if whitening was applied prior to learning the mapping;
\item De-whitening to capture the variance of the original embedding spaces, applied only if whitening was applied prior to learning the mapping; and
\item Reducing the dimension by only keeping the most important components of the source and target spaces, helping to remove noise captured in the tail components.
\end{itemize}

The authors show that combining these steps helped them achieve superior performance when using CBOW word vectors and 5,000 supervised training examples.
The full framework has been packaged and released as open source code under the moniker VecMap, and is used as a baseline in many comparative surveys.

\paragraph{Margin Models}

The methods of the prior two sections rely on variations of mean-squared error to compute and learn from the differences between the source and target space.
An alternative modeling technique leverages a max-margin based loss function.
These objectives seek to reward the weights associated with positive pairs (in this case, words that are direct translations) while reducing the signal from noise pairs generated either randomly or using a heuristic.
In the case of word-to-word translation, the association between pairs is defined by their cosine similarity, thus we may define the max-margin loss as 
\begin{center}
$$ \sum_{i=1}^{n} \sum_{j \neq j}^k \max \{ 0, \gamma - \cos(Wf_1(w_i^s), w_i^t) + \cos(Wf_1(w_i^s), w_j^t) \} $$
\end{center}
where $k$ represents the number of noise pairs (negative samples) and $\gamma$ is a parameter fixed for setting the margin between positive and negative cases.
Using this objective was first proposed by~\cite{margin-hubness} to address issues of hubness seen in regression and orthogonal techniques. 
The presence of hubs is driven by embeddings that dominate the space due to their high cosine similarity with all other vectors in the source or target space.
These hubs can be caused by the overall frequencies of words in the underlying corpus~\cite{frequency-agnostic}, a common mean vector present in all word vectors causing issues of anisotropy in the embedding spaces~ \cite{but-the-top}, or issues derived from least-squares regression where low variance points are all grouped together in the target space.

Margin-based models are also explored in~\cite{margin-retrieval}, where the authors also aim to combat the issue of hubs by introducing a new retrieval criteria.
The cross-domain similarity local scaling (CSLS) is defined as
\begin{center}
$$ CSLS(x,y) = -2\cos(x,y) + \frac{1}{k} \sum_{y' \in N_Y(x)} \cos(x, y') + \frac{1}{k} \sum_{x' \in N_X(y)}\cos(x', y)$$
\end{center}
where $N_Y(x)$ is the set of $k$ nearest neighbors of $x$ in the target space. 
The authors build this retrieval criteria into their margin model by using unpaired words (those with no explicit translation in the training set) as negative samples when computing nearest neighbors.
The full objective function, called the relaxed CSLS (RCSLS) is then computed as 
\begin{center}
$$ \frac{1}{n} \sum_{i=1}^n -2\cos(Wf_1(w_i^s), w_i^t) + \frac{1}{k} \sum_{w_j \in N_Y(Wf_1(w_i^s))} \cos(Wf_1(w_i^s), w_j) + \frac{1}{k} \sum_{Wf_1(w_j^s)\in N_X (w_i) } \cos(Wf_1(w_j^s), w_i) $$
\end{center}
For margin-based methods, RCSLS is the most popular method, used as a benchmark for comparisons to other methods~\cite{proper-eval, lost-evaluation}.
While competitive, we find margin-based methods are studied less frequently; we conjecture this is due to the difficulty in selecting informative negative samples and the preference for methods using as little supervision as possible.

\paragraph{Other Approaches}

In the interest of exploring methods that extend past word-to-word alignment and are able to generalize to other embedding spaces, we briefly mention alignment methods that lie outside the three categories noted above.
The first such method relies on the word neighborhood structures, based on the assumption that for a neighborhood of points in the source space the neighborhood can be reconstructed after applying a linear mapping to the target space.
By using manifold learning, without making the assumption that the manifolds (or embedding spaces) were learned via the same algorithm, to capture neighborhood structures,~\cite{locality-preserving} propose a new locality preserving loss function.
Given embedding spaces as manifolds $M^s$ and $M^t$, the goal is to learn a mapping $f: M^s \rightarrow M^t$, which is optimized based on three pieces: an orthogonal piece; a mean-squared error piece; and an additional locality preserving loss piece.
This approach represents the base of several models, encapsulating the regression class of models and the orthogonal constrained class of models while adding structure preserving pieces for regularization.
The entire loss function can be written as 
\begin{center}
$$\mathcal{L} = \mathcal{L}_{mse}(\theta_f) + \beta \mathcal{L}_{lpl}(\theta_f, W) + \mathcal{L}_{orth}(W)$$
\end{center}
where $\mathcal{L}_{mse}$ and $\mathcal{L}_{orth}$ are as defined in prior sections and 
\begin{center}
$$\mathcal{L}_{lpl}^i = \norm{m_i^t - \sum_{m_j^s \in N_k(m_i^s)} W_{ij} \cdot f(m_j^s)}^2   $$
\end{center}
with $\beta$ a constant to control the influence of the LPL loss and $m_q^s \in M^s, M_q^t \in M^t$.
The locality preserving loss was shows to assist in both sentence space and word space alignment, particularly when training dataset sizes are limited.

\subsubsection{Semi-Supervised Methods}\label{sec:word-semi}

In instances where full parallel corpora or dictionaries are not readily available it is possible to use smaller seed lexicons to build toward larger datasets in a semi-supervised way.
One such way of building pseudo-dictionaries' is to identify words that are expressed as the same string in both the source and target language~\cite{inverted-softmax}.
These typically occur for proper nouns and abbreviations such as FBI and Microsoft. 
According to~\cite{inverted-softmax}, this procedure was able to generate nearly 47,000 translation pairs between English and Italian, much larger than the 5,000 most popular terms used by many supervised methods, with excellent evaluated levels of precision.

Aside from string matching, other alternative corpus construction methods include using very small seed lexicons (on the order of 25 word pairs) and iteratively adding candidate pairs.
The work of~\cite{biwe} propose alternating between a step for learning the mapping $W$ similar to those in Section~\ref{sec:orth-word}, followed by a dictionary induction step.
Given embedding spaces  $X$ and $Z$, let $D$ be the binary matrix representing the word-pair dictionary between the two languages, i.e. $D_{ij}=1$ when word $i$ of the source is aligned to word $j$ in the target language. 
The mapping matrix can then be defined as 
\begin{center}
$$ W^{*} = \argmin_{W} \sum_{i} \sum_{j} D_{ij} \norm{WX - Z}^2 $$
\end{center}
At each step of processing, updates to $D$ are computed as $D_{ij} = 1$ if
\begin{center}
$$ j = \argmax_{k} (X_{i*}W^{*}) \cdot Z_{k*} $$
\end{center}
otherwise $D_{ij} = 0$.
To evaluate the efficacy of this method, small seed dictionaries are sampled ranging in size from 25 to 2,500 entries, as well as experimenting with only aligning numerals (i.e. digits 0 to 9).
Given the limited training set, this self-learning paradigm is competitive with, and at times outperforms, supervised methods.

The work of~\cite{bilingual-non-iso} further explores iterative learning, alternating between supervised alignment and unsupervised distribution matching, as explored in the next section, as well as introducing novel metrics to assess the orthogonality assumptions used in supervised approaches.
We further unpack these notions in Section~\ref{sec:word-unsuper}.

\subsubsection{Unsupervised Methods}\label{sec:word-unsuper}

Under the goal of restricting the amount of parallel data needed to create an alignment between two word spaces, several approaches have been proposed that attempt to leverage the structure of the embedding space itself, completely removing the need for parallel data.
A key approach was described in~\cite{non-parallel-conneau} where the authors propose leveraging an adversarial learning paradigm.
In this setup, the goal is still to learn a linear map $W$ between the source embedding vectors $f_1(w_i^s)$ and target space embeddings $f_2(w_i^t)$.
A discriminator $D$ is trained to recognize and separate the mapped embeddings $W f_1(w_i^s)$ from $f_2(w_i^t)$, while an adversarial generator $G$ is trained to fool $D$.
The given loss functions for both models are
\begin{center}
$$ \mathcal{L}(\theta_D | W) = -\frac{1}{n} \sum_{i=1}^n \log P_{\theta_D}(source = 1| Wx_i) - \frac{1}{m} \sum_{i=1}^m \log P_{\theta_D}(source = 0 | y_i)$$
\end{center}
for the discriminator model, and
\begin{center}
$$ \mathcal{L}(W | \theta_D) = -\frac{1}{n} \sum_{i=1}^n \log P_{\theta_D}(source = 0| Wx_i) - \frac{1}{m} \sum_{i=1}^m \log P_{\theta_D}(source = 1 | y_i)$$
\end{center}
for the generator model.
With an initial linear map $W$ learned, the authors then apply a refinement procedure by identifying anchor points as pairs that were frequently identified as translations in the prior step.
The anchor points and their corresponding word frequencies are used to solve the orthogonal Procrustes problem to generate a refined mapping matrix $W^*$.
This final matrix is used in conjunction with the CSLS objective described in prior sections to mitigate hubness and areas of density in generating a final translation from source to target.
The adversarial method has been utilized to generate large benchmark datasets under the name Multilingual Unsupervised or Supervised Embeddings (MUSE), releasing parallel embedding spaces trained using FastText in 110 languages. 

As previously discussed, word embedding models tend to reflect the frequency of word usage in the underlying language.
While the adversarial method directly leverages word frequencies, an alternative unsupervised method in~\cite{self-learning} captures these patterns by analyzing the similarity distributions of the word vectors themselves.
By constructing a pair-wise similarity matrix of all word embeddings in the source and target languages, trends in their usages can be exploited to create an initial seed dictionary.
By pre-processing to unit normalize the embeddings in both the source $X$ and target $Z$ spaces, these similarity matrices can quickly be computed as $M_X = XX^\top$ and $M_Y = YY^\top$.
To further reduce the complexity of finding maps between these similarity matrices, each similarity matrix can then be sorted row by row to identify the most influential embedding dimension and nearest neighbor searches can then be executed to generate candidate pairs.
The seed dictionary can then be expanded using semi-supervised methods described in~\cite{biwe}. 

Aside from leveraging similarity distributions of the underlying embedding spaces, methods are also available to treat these embedding spaces as metric spaces, adopting mathematical tools from measure theory and topology to describe their nature.
One such metric is the Gromov--Wasserstein distance used to compare two pairs of spaces, rather than the pairwise point-by-point metrics such as similarities. 
Using this metric,~\cite{gw-align} transform the alignment problem to one of finding an optimal transport from source $X$ to target $Z$.
Due to computational costs, the problem is split into two steps where the two spaces are first aligned using the explicit optimization to find an optimal coupling followed by a refinement using an orthogonal Procrustes procedure, as in~\cite{non-parallel-conneau}.

\subsection{Graph-to-Graph Alignment Techniques}\label{sec:graph-to-graph}

As in word-to-word alignments, graph-to-graph alignment techniques can be classified into supervised, semi-supervised and unsupervised methods.
Within each paradigm, however, it is slightly more complicated to develop a straight-forward classification of techniques.
We posit this is due not only to the variety of graph datasets available but the velocity at which new research is being published, as noted in Section~\ref{sec:landscape}.
We proceed by categorizing techniques by their level of parallel data needed to learn a robust model.
Where applicable, we will also classify techniques according to their approach to the source and target graph embeddings, noting if they utilize translation-based, semantic-matching or graph-structure models.

\subsubsection{Supervised Methods}

To address issues of coverage in cross-lingual knowledge graphs,~\cite{mtranse} propose a method for embedding knowledge graphs in a source and target language and automatically learning alignments between them, called MTransE.
Leveraging the translational-based TransE algorithm for generating embeddings of each monolingual knowledge graph $G_i$ and $G_j$, the embedded triples of each graph are then fed through an alignment scoring function $S_a$, where the total alignment score is calculated as
\begin{center}
$$ S_A = \sum_{(T, T') \in \delta(G_i, G_J)} S_a(T, T')  $$
\end{center}
where $\delta_{(G_i, G_j)}$ represents the supervised set of pre-aligned triples.
The authors propose three main classes of functions for $S_a$: distance-based measures, translation vectors and linear transformations.
For distance-based measures, the triples in $G_i$ and $G_j$ can be represented as a function of the difference in the head and tail entities
\begin{center}
$$ S_{a_1} = \norm{h - h'} + \norm{t - t'} $$ 
\end{center}
or adjusted to also represent differences in the relation embeddings
\begin{center}
$$ S_{a_2} = \norm{h - h'} + \norm{r - r'} + \norm{t - t'} $$
\end{center}
An alternative approach focuses not only on the differences in individual components of the triples, but allows for translation vectors to be learned between the entities and relations, as defined by
\begin{center}
$$ S_{a_3} = \norm{h + v_{ij}^e - h'} + \norm{r + v_{ij}^r - r'} + \norm{t + v_{ij}^e - t'} $$
\end{center}
where $+ v_{ij}^e$ and $+ v_{ij}^r$ are learnable translations such that $ e + v_{ij}^e \approx e'$.
The final class of functions defines learnable linear transform matrices, with one focused on translations between entities
\begin{center}
$$ S_{a_4} = \norm{M_{ij}^e h - h'} + \norm{M_{ij}^e t - t'} $$
\end{center}
and another with learnable transformations for both entities and relations
\begin{center}
$$ S_{a_5} = \norm{M_{ij}^e h - h'} + \norm{M_{ij}^r r - r'} + \norm{M_{ij}^e t - t'} $$
\end{center}
The authors conclude that the linear transformation models work best with limited differences between the model with entity transformations and the model with both entity and relational translations.

The methods of MTransE lean heavily on the underlying translation model of TransE, which, while directly addressing similarities in graph structures in each individual space, largely ignores other information sources contained in the knowledge graph such as entity types and attributes. 
Relying only on the structured embedding approaches also leads to issues when the distribution of relations in the knowledge graph is skewed, as has been widely observed in many large-scale knowledge graphs~\cite{oneshot, fewshot}.
To leverage both structure and attributes in aligning both graphs, a joint attribute-preserving embedding (JAPE) module is presented in~\cite{jape}.
For the structure embedding piece, the authors again leverage a translation-based approach letting $f(p) = \norm{h + r -t}^2$ where $p$ is a known triple from the supervised training set.
They then slightly modify the training process by using a training set $P$ that has pre-aligned triples in both the source and target space to capture correspondences between entities sharing similar relationships.
This accomplishes the alignment of both cross-lingual entities and their relations in a single computational step and can be seen as optimizing the score of
\begin{center}
$$ \mathcal{L_{SE}} = \sum_{p \in P} \sum_{p' \in P'} (f(p) - \alpha f(p')) $$ 
\end{center} 
where $\alpha$ is a tunable margin hyper-parameter and $P'$ a set of negative samples.
In addition, attributes of the entities such as their data type and correlations between relation occurrences are used to generate attribute embedding vectors using a skip-gram like objective function
\begin{center}
$$ \mathcal{L_{AE}} = -\sum_{(a,c) \in H} w_{a,c}\cdot \log p(c|a) $$
\end{center}
where $H$ is the set of pairwise positively correlated attributes, i.e. when entity $e_j$ has attribute $a$ it is also highly likely to have attribute $c$.
The attribute embeddings are then used to compute three similarity matrices $S^(1), S^(2), S^{(1,2)}$ representing the inner-graph entity attribute similarity scores as well as the cross-graph attribute similarity scores. 
This additional data from the training set helps to build more support for entities to be aligned that can not be captured when using only translational-based models and can be combined to minimize the objective
\begin{center}
$$ \mathcal{L_S} = \norm{ E_{SE}^{(1)} - S^{(1,2)} E_{SE}^{(2)}} + \beta ( \norm{E_{SE}^{(1)} - S^{(1)} E_{SE}^{(1)}} + \norm{E_{SE}^{(2)} - S^{(2)} E_{SE}^{(2)}} ) $$
\end{center}

The structured and attribute losses can then be jointly optimized for learning the entire model through
\begin{center}
$$ \mathcal{O_{JOINT}} = \mathcal{O_{SE}} + \delta \mathcal{O_{S}}  $$
\end{center}
where $\delta$ is a tunable hyper-parameter to moderate the influence of the attribute similarities.

Having demonstrated the importance of incorporating both structure and attributes into the alignment process, other authors followed in the footsteps of the JAPE model, although with different underlying embedding techniques.
In the work of~\cite{gcn-ea}, the authors utilize the graph convolutional network architecture (GCN-EA) to embed entities from the training sets into an aligned space.
By creating a bipartite graph between the pre-aligned entities in the training set the GCN-EA approach models the edges between each distinct graph as equivalence relations, discovering other equivalence relations as alignments by encoding neighborhood information.
This approach is then further refined by incorporating the entity attributes in an additional convolutional layer after entity embeddings have been defined.
Given two graphs $G_1$ and $G_2$ and a training dataset of pairs of matched entities from each, i.e. $S={(e_{m_1}, e_{m_2})}, e_{m_1} \in G_1, e_{m_2} \in G_2$, we define two parallel GCN models $GCN_1$ and $GCN_2$ to generate embeddings of each input graph.
Each GCN outputs a vector representation of a given input entity, call them $v_{m_1}$ and $v_{m_2}$ for $e_{m_1}$ and $ e_{m_2}$, that can be seen as the concatenation of two parts.
The first piece of the output vector $v_{m_j}$ represents the structural piece from the convolutional network with dimension $d_s$.
The second piece represents the attribute representation embedding from the next layer of the network, with dimension $d_a$.
Thus each vector $v_{m_j}$ is of $d_s + d_a$ dimension and compactly represent the structure and attributes of the particular input entity.
These representations are then fed to a distance matching function
\begin{center}
    $$ D(x,y) = \beta \frac{f(h_s(x), h_s(y)}{d_s} + (1 - \beta) \frac{f(h_a(x), h_a(y)}{d_a} $$
\end{center}
where $f(a,b) = \norm{a-b}$, $h_s$ and $h_a$ take the structure and attribute piece of the embedding, respectively, and $\beta$ is a hyperparameter that balances the trade-off between the importance of structure and attributes.
The distances between pre-aligned entities in the training set can then be back-propagated through the network using a margin-based criteria, one for the structure embeddings and one for the attribute embeddings
\begin{center}
$$ \mathcal{L}_s = \sum_{(x,y) \in S }\sum_{(x',y') \in S'} [f(h_s(x), h_s(y)) + \gamma_s - f(h_s(x'), h_s(y'))]_{+} $$

$$ \mathcal{L}_a = \sum_{(x,y) \in S} \sum_{(x',y') \in S'} [f(h_a(x), h_a(y)) + \gamma_a - f(h_a(x'), h_a(y'))]_{+} $$
\end{center}
with $\gamma_s$ and $\gamma_a$ as margin hyper-parameters.

Thus far, the three KG-to-KG alignment methods explored, namely MTransE, JAPE and GCN-EA, have focused only on the problem of aligning entities between the two input graphs, and while attribute information from the graphs has also been included, little has been done to also factor in the relations and relational-types in each individual graph.
Incorporating relation information is an important facet for selecting an approach to aligning an input source to a knowledge graph embedding, especially in the application to relation extraction from text documents.
Equally important in capturing relational data is accounting for directionality; methods must be able to distinguish between one-to-one, many-to-one and many-to-many relation types.  
Rather than solely relying on aligned entities,~\cite{mmea} create a training set of aligned entities $S_e={(e_{m_1}, e_{m_2})}, e_{m_1} \in G_1, e_{m_2} \in G_2$, and aligned relations $S_r={(r_{m_1}, r_{m_2})}, r_{m_1} \in G_1, r_{m_2} \in G_2$ in a multi-mapping relation aware technique dubbed MMEA.
The authors proceed by defining their own knowledge graph embedding process, avoiding the pitfalls of translation-based methods by defining their own embedding process, called DistMA.
DistMA works by replacing translation-based distances with inner-products, defined by
\begin{center}
$$ E_1(h,r,t) = \langle v_h, v_r \rangle + \langle v_r, v_t \rangle + \langle v_h, v_t \rangle $$
\end{center}
and replace the margin-based optimization with a logistic loss function, defined by
\begin{center}
$$ -\sum_{(h,r,t) \in S^+} \log \sigma(E_1(h,r,t)) - \sum_{(h',r',t') \in S^-} \log \sigma(-1 \cdot E_1(h',r',t')) + \lambda \norm{\theta} $$ 
\end{center}
Using the inner product rather than subtraction-based distances allows the embeddings to scale well to multi-relational facts in the graph, where methods like TransE typically struggle. 
The downside is that the proposed DistMA is highly symmetric, making no distinction between the head and tail of the triple, thus incorporating no sense of the directionality of the relation.
To combat this, the authors also leverage the ComplEx embedding method, previously presented in Section~\ref{sec:semantic-matching}.
Letting
\begin{center}
$$ E_2 =  \Re(\langle w_h, w_r, w_t \rangle), w_i \in \mathbb{C}$$
\end{center}
the final scoring function for each triple is combined and written as
\begin{center}
$$ E(h,r,t) = E_1(h,r,t) + E_2(h,r,t) $$
\end{center}
With both entities and relations embedded for each knowledge graph, the airs from the training set are aligned in a common embedding space such that their representations are equal. 
This is accomplished by using a cosine similarity metric
\begin{center}
$$ sim(e_i, e_j) = \langle \norm{v_{e_i}}, \norm{v_{e_j}} \rangle $$
\end{center}
to build a similarity matrix$S_{1,2}$ between the two knowledge graphs. 
This similarity matrix can then be ranked from both directions, i.e. for the similarities $M_{1,2}:  G_1 \rightarrow G_2$ and $M_{2,1}: G_2 \rightarrow G_1$.
The final ranking matrix is then computed as $M = M_{1,2} + M_{2,1}^\top$.

In continuation of research on addressing multi-relational patterns,~\cite{ntam} focus on Non-Translational Alignment for Multi-relational networks (NTAM). 
Rather than relying on a semantic energy, translational-based, or graph convolutional model to build embeddings, the authors build a probabilistic model based on \textit{motifs} that can be found within the graph.
These motifs, or graph patterns, include triangular structures in the graph where a given node in the triangle can have in-degree of zero (out-degree two), one (out-degree one) or two (out-degree zero), as is accomplished in~\cite{structural-prob}.
While these motifs are flexible in capturing local structures in the graph, aligning these structures required nodes in each individual graph to exhibit very similar neighborhood structures, an assumption that may not hold in large-scale heterogeneous graphs.

For the majority of knowledge graph alignment approaches discussed above, the underlying embedding algorithms rely on negative samples, or false facts, to be generated.
These negative sampling paradigms inherently make use of the closed-world assumption wherein all facts are assumed to be contained in the knowledge graph.
In opposition is the open-world assumption, where we have only build a knowledge graph of our currently known facts, and the validity of those not contained in this set is uncertain.
Using negative sampling instantiates a closed-world assumption, and when those negative facts turn out to actually be true, model performance suffers.
In opposition, adversarial networks leverage two networks that attempt to trick one another. 
The first network, the generator, attempts to create samples that look similar to those in the original data distribution, yet are created in a synthetic way.
The job of the second network, the discriminator, is to differentiate between instances of the true dataset versus those coming from the generator.
Adversarial networks have been utilized to generate embeddings of single knowledge graphs~\cite{kbgan} and can also be used in aligning the representations of distinct knowledge graphs.
Based on the notion that the embedding spaces of each graph should have similar spatial features for entities that are likely the same, called by the authors of~\cite{ake} the embedding distribution, an adversarial network can be used to learn to discriminate between these embedding distributions in order to learn an approximate isomorphism between the two spaces.
To accomplish this task, the authors introduce the representation module, the mapping module and the adversarial module.
In the representation module, two separate instances of the TransE model are trained on each graph, creating two sets of embeddings $e_s$ and $e_t$.
These embedding matrices are then fed into the mapping module, where seed pairs $S=(e_{s_i}, e_{t_i})$ are used to learn a linear mapping, defining the loss function as 
\begin{center}
$$ L_M = \sum_{(e_s, e_t) \in S} \norm{G e_s - e_t}$$
\end{center}
As in approaches for word-to-word embedding alignment, $G$ can be restricted to be an orthogonal matrix.
The authors introduce two additional constraints: the feature reconstruction constraint and the mapping reconstruction constraint.
Intuitively, the feature reconstruction constraint dictates that once an embedding for an entity is mapped from the source space to the target space, that same mapping can be applied to map it back to the source space representation.
This can be reflected in the adjusted loss function
\begin{center}
$$ L'_M = \sum_{(e_s, e_t) \in S}  \lambda_1 \norm{Ge_s - e_t} + \lambda_2 (\mu \norm{e_s - G^\top G e_s} + (1-\mu) \norm{e_t - G G^\top e_t})$$
\end{center}
where $\lambda_1 , \lambda_2$ are learnable weights to balance the reconstruction constraint and $\mu$ is a harmonic factor.
The mapping reconstruction constraint is a variant on restricting $G$ to be orthogonal, forcing the learning algorithm to push $G$ toward the nearest orthogonal manifold, modifying the loss function as
\begin{center}
$$ L''_M =  \sum_{(e_s, e_t) \in S} \lambda_1 \norm{Ge_s - e_t} + \lambda_2 \norm{G^\top G - E}_F$$
\end{center}
where $E$ is the identity matrix and $F$ is the Frobenius norm.
With a mapping learned, the mapped embeddings can be fed to the adversarial module where their synthetic counterparts are build by a generator network while the discriminator network learns to differentiate between true and false examples.
The authors show that the adversarial setup helps with generalization as its main focus is on aligning the topological features of each embedding space in a way which reduces sensitivity to noise.

\subsubsection{Semi-Supervised Methods}

While the issue of building supervised word-to-word datasets is a challenge for word alignment techniques, the issue is even more prevalent in knowledge graphs due to their large degree of heterogeneity. 
Building links between entities in disparate knowledge graphs often requires the intervention of human experts and comes at a significant cost.
Rather than rely solely on labeled instances, semi-supervised approaches build from a set of seed aligned entities, iteratively building confidence in newly aligned pairs and expanding the set of training examples.

By embedding the separate graphs using translational-based methods, the authors of~\cite{iptranse} build a joint embedding space and utilize a soft alignment scoring function to estimate a reliability score of aligned entities.
These three modules are trained in an iterative fashion, making updates to the training set and the joint embeddings at each step.
For the individual embeddings, the authors utilize the path-inclusive embeddings of PTransE~\cite{ptranse}, generating two entity embedding sets $E_1$ and $E_2$.
To build a joint embedding space, the authors propose three methods: a translation-based model, a linear model and a parameter sharing model.
The translation-based model, IPTransE, introduces a new alignment relation $r$ that maps $e_s \in E_1$ to $e_t \in E_2$, where
\begin{center}
$$ E(e_s, e_t) = \norm{e_s + r^{(E_1 \rightarrow E_2)} - e_2}$$
\end{center}
The linear mapping model replaces this relation with a transformation matrix $M$ such that
\begin{center}
$$ E(e_s, e_t) = \norm{M^{(E_1 \rightarrow E_2)}e_1 - e_2} $$
\end{center}
The parameter sharing attempts to make no mapping, instead forcing aligned entities from the seed set $\mathbb{L}$ to have the same embedding representation, such that
\begin{center}
$$ e_s \equiv e_t, (e_s, e_t) \in \mathbb{L}$$
\end{center}
Once entities are projected into a joint space, each entity embedding in the source space is compared to all entity embeddings in the target space, building an aligned entity where
\begin{center}
$$ \hat{e_t} = \argmin(E(e_s, e_t)), E(e_s, \hat{e_t}) < \theta$$
\end{center}
where $\theta$ is a hyperparameter controlling the distance. 
The pair $(e_s, \hat{e_t})$ can then be added to $\mathbb{L}$ and the joint alignments can be updated accordingly. 
In addition to simply updating the set $\mathbb{L}$, the authors also introduce a soft alignment function
\begin{center}
$$ R(e_s, e_t) = \sigma(k(\theta - E(e_s, e_t)) $$
\end{center}
that tracks the reliability of the new pair, where $\sigma$ is the softmax function and $k$ is a tunable hyperparameter.

Also leveraging translational-based models for embedding each knowledge graph, semi-supervised entity alignment with degree differences (SEA)~\cite{sea} adjusts the TransE approach by incorporating information about each entities degree when building the embeddings. 
The key insight is that entities that are well-connected appear in more triples, and thus have more robust embeddings containing more information than entities that are infrequently occurring in the graph.
The more frequently occurring entities thus form hubs in each embedding space, making the alignment maps learned biased toward these points.
Adjusting TransE to better reflect the degree distribution of each entity helps to alleviate these issues and build more robust alignment maps.
To prevent entities with similar degree from clustering together in the embedding space,  the authors use an adversarial network where a generator builds degree-aware embeddings while two discriminators $D_1$ and $D_2$ are used to classify entities with high or normal degree and entities with low degree, respectively.
By designing the generator to create high-quality embeddings to fool the discriminators, those embeddings will encode the entities degree in such a way that embeddings of various degrees become linearly inseparable, and thus don't occupy a dense area of the embedding space.
The adversarial training is done by training the TransE representations with the discriminators fixed, then alternating by training the discriminators with the embeddings fixed, generating two sets of degree-aware KG embeddings $\theta_i^1$ and $\theta_j^2$.
To align the entities from the set of pre-labeled seed alignments $\mathbb{L}$, cycle consistent translation matrices are learned by minimizing
\begin{center}
$$ \sum_{(e_i, e_j) \in \mathbb{L}} \norm{M^1 \theta_{e_i}^1 - \theta^2_{e_j}} + \norm{M^2\theta_{e_j}^2 - \theta^1_{e_i}}$$
\end{center}
where the cycles 
\begin{center}
$$  \theta_{e_i}^1 \rightarrow M^1 \theta_{e_i}^1 \rightarrow M^2 M^1 \theta_{e_i}^1$$ 
$$  \theta_{e_j}^2 \rightarrow M^2 \theta_{e_j}^2 \rightarrow M^1 M^2 \theta_{e_j}^2$$ 
\end{center}
help to improve generalizability to unlabeled instances. 
By using all the unlabeled entities in generating the degree aware embeddings, the SEA model is able to leverage both labeled and unlabeled entities in building a robust alignment.

By incorporating a bootstrapping approach, the authors of~\cite{bootea} completely abandon the translational-based model and opt instead for a margin-based model designed to directly leverage information contained in the positive and negative samples sets, which are continuously expanded as the model trains.
For triples $\tau$, the objective function
\begin{center}
$$ \mathcal{O}_e = \sum_{\tau \in \mathbb{T}^+} [f(\tau) = \gamma_1] + \mu_1 \sum_{\tau' \ in \mathbb{T}^-} [\gamma_2 - f(\tau')]$$
\end{center}
where $\mathbb{T}^+, \mathbb{T}^-$ refer to the sets of positive and negative triples, respectively.
In building $\mathbb{T}^+$ and $\mathbb{T}^-$, the authors introduce an $\epsilon$-truncated negative sampling strategy, emphasizing that corruption of the head or tail entity should be done so in an intelligent way to maximize the signal the model can learn from.
In this negative sampling paradigm, the negative candidates are selected from a neighborhood of $s=\ceil[\big]{(1-\epsilon)N}$ based on the cosine similarity of their embeddings.

After $t$ steps of training, the set $\mathbb{T}^+$ is updated based on the bootstrapping procedure.
As these new bootstrapped instances may contain errors, the authors introduce an editing technique to dampen their effect.
Prior to new candidates $y$ and $y'$ with a truth label $x$ being added to $\mathbb{T}^+$, they are evaluated through
\begin{center}
$$ \Delta_{(x,y,y')}^{(t)} = \pi(y|x; \Theta^{(t)}) - \pi(y'|x; \Theta^{(t)}) $$
\end{center} 
to determine the highest likelihood of the label, preventing uncertainty from leaking into the bootstrapped training set.

\subsubsection{Unsupervised Methods}

While there is limited research in semi-supervised methods for knowledge graph alignment, fully unsupervised methods are even less common.
In their survey of the literature, ~\cite{kg-align-survey} claim to observe no research articles on unsupervised methods for knowledge graph alignment.
In the time between the release of the survey and this publication, we have found an example of authors exploring unsupervised techniques.
In~\cite{unsuper-graph-align} an adversarial training paradigm is used to build links between two graphs.
The embeddings of each graph are generated using the DeepWalk technique, taking advantage of the structural properties of each individual graph.
These are then mapped using a matrix $W$ fed to a discriminator to differentiate between the source and target space.
Given the recency of this publication, its current lack of peer review, and experimentation using only social network domains, we leave the other details to the reader but include its mention to highlight an area of growing interest for researchers.

\subsection{Sentence-to-Sentence Alignment Techniques}\label{sec:sentence-align}

While many of the techniques applied to word-to-word alignment also apply to sentences, there is also a line of research that focuses only on techniques for sentence alignment.
Sentence alignment introduces an additional complexity over word alignment due to variability in word ordering and syntactic and morphological differences between languages that challenge the efficacy of traditional mapping based systems. 
Applications in this space tend to focus on applications to neural machine translation (NMT), however, we believe that these techniques have applications to other research domains.

\subsubsection{Supervised Methods}

While there are a limited number of publications exploring supervised sentence to sentence embedding alignment, there are applications of existing word-to-word techniques to this domain. 
In~\cite{absent}, to benchmark their semi-supervised method the authors re-implement several alignment models for the purpose of mapping sentence embeddings for machine translation.
Specifically, they use the linear regression model model from~\cite{exploit-sim}, the $l_2$ regularized model of ~\cite{hubness}, and the inverted softmax model of~\cite{inverted-softmax}.
Details of each of these approaches are given in Section~\ref{sec:word-supervised}.
In their evaluation of these techniques, the authors find that these simple linear techniques and their extensions outperform the more advanced models from seq2seq~\cite{seq2seq}, fairseq~\cite{fairseq} and LASER~\cite{laser}.
As these more advanced models do not perform explicit alignments, we leave these for the reader to explore.

\subsubsection{Semi-Supervised Methods}

With an eye toward reducing the amount of parallel data necessary,~\cite{absent} utilize bidirectional GANs for aligning sentence representations.
In addition to defining a piece of the loss function for sentence representation pairs $(x,y)$ that are in the training set, the authors also use all available non-parallel data in their approach.
The resulting objective function 
\begin{center}
$$ \mathcal{L}_{real} = \mathbb{E}_{x,y}[\log(D_{real}(x,y)] + \mathbb{E}_{x}[\log(1-D_{real}(x, G_{X}(x)))] + \mathbb{E}_{y}[\log(1-D_{real}(y, G_{Y}(y)))] $$
\end{center}
where a real pair $(x,y)$ in the parallel set is contrasted with fake pairs $(x, G_X(x))$ and $(y, G_Y(y))$ created by the respective generators for the source and target space.
To further leverage the data available from the non-parallel sentences, the authors additionally introduce two loss functions.
The first is designed to minimize the expected value of mismatched pairs, or negative samples $(x', y') \in X \times Y$ such that
\begin{center}
$$ \mathcal{L}_{mis} = \mathbb{E}_{x',y'}[\log(1-D_{real}(x',y'))] $$
\end{center}
The second loss function added includes an additional discriminator to distinguish whether the sentence embedding came from the source or target space, defined as
\begin{center}
$$ \mathcal{L}_{dom} = \mathbb{E}_{x}[log(D_{dom}(x, G_{X}(x)))] + \mathbb{E}_{y}[log(1 - D_{dom}(y, G_{Y}(y)))] $$
\end{center}
These three loss functions contribute equally to the overall model loss
\begin{center}
$$ \mathcal{L} = \mathcal{L}_{real} + \mathcal{L}_{mis} + \mathcal{L}_{dom}$$
\end{center}

To represent the sentences in both the source and target space, the authors use FastText word vectors and simply average each word representation to create a sentence embedding.
In their ablation study, the authors additionally experiment with TF-IDF weighting, finding that for corpora with longer sentence length the TF-IDF weighting leads to improved accuracy, while shorter sentences do not exhibit similar gains. 
The intuition behind these findings is that much of the noise in longer sentences, such as stop words and other semantically irrelevant words used for syntactic purposes, can be down-weighted, thus creating a more semantically meaningful sentence representation.

\subsubsection{Unsupervised Methods}

Due to the added complexities of mapping full sentence representations, there is limited research on completely unsupervised sentence embedding mappings.
For that reason, we restrict our evaluation to the methods demonstrated in \cite{intersemrep}.
To address the issue of lack of parallel corpora for supervised alignment, the authors of \cite{intersemrep} introduce the notion of interlingual semantic representations (ISR) for the few or zero-shot cases.
ISR attempts to create an intermediate, low-dimensional space that captures word and sentence semantics that can be fine-tuned to any language or downstream task.
To accomplish this representation, the authors utilize an adversarial approach, building a sequence of generators and discriminators to encode language into intermediate representations using only a monolingual corpus.
A single generator $G$ makes use of an encoding step \textit{enc} to map an input sentence $s$ to ISR, then applies a decoding step \textit{dec} back to a translated form of that sentence $\hat{s}$.
The translation is then fed to the discriminator $D$ for the dual task of determining if the sentence translation is real or synthetic, as well as making a classification for the language of that translation.
Simultaneously, the translation $\hat{s}$ is backpropagated through the generator $G$ for calculation of two losses.
The first loss, called the ISR loss, minimizes the distance between the ISR of the forward translation and the backward translation, helping to fine tune the intermediate embedding layer.
The second loss, the reconstruction loss, measures how closely the original input sentence vector $s$ and the forward-backward representation after two applications of the generator $G(G(s))$ match.
This type of cycle-consistency loss has been shown to aid in cross-domain translation, as in \cite{CycleGAN2017}, and the authors claim that cycle-consistency helps to preserve semantic information flow from the source to intermediate representations.
For fixed embedding inputs, the authors use a BERT-as-a-service model for generation, train their neural architecture using the XNLI dataset, and demonstrate through an ablation study the importance of the reconstruction loss, showing that the cycle-consistency constraint does help in the zero-shot learning task. 
Fully unsupervised sentence translation remains a challenging yet open research problem, and the pace of research publications in this area continues to increase.

\section{Benchmark Datasets}
\label{sec:benchmarks}
In this section, we document some of the most popular datasets used in the alignment literature.

\subsection{Cross-language Word Alignment Benchmarks}

As introduced in Section~\ref{sec:language-translation}, the task of Bilingual Lexical Induction aims to evaluate the consistency and ability to learn alignments between embeddings of two distinct languages.
There are two themes that datasets in this space may be classified as: those that provide aligned source and target text (akin to datasets used for machine translation) and those that provide pre-trained embeddings of the source and target languages along with seed alignments.
In regards to the latter, such published datasets typically select a base embedding algorithm (i.e.\ FastText or GloVe) and generate pre-trained embeddings on several monolingual corpora, each using the same model hyper-parameters to dictate consistency. 

A modern and oft-cited toolkit for BLI datasets is the Facebook MUSE dataset~\cite{non-parallel-conneau}. 
This dataset contains monolingual embeddings and seed dictionaries for 30 languages, as well as bilingual seed dictionary pairs for 110 languages.
For languages coupled with monolingual embeddings, all such embeddings were generated using the FastText algorithm trained over a copy of Wikipedia in the respective language.
Each set of embeddings has been generated using the Skip-gram model with the embedding dimension set to 300 with no additional parameter tuning.
While this paradigm allows for consistency in comparing the embedding spaces, it could be the case that the embeddings may perform better on certain languages, confounding the evaluation of the structural similarities between embedding spaces.
For the ground-truth seed dictionary pairs, aligned seeds are provided in sets of 5,000 training pairs and 1,500 testing pairs. 
This benchmark was further criticized in~\cite{lost-evaluation} due to the large presence of proper nouns, which are considered to be only referential and contain limited lexical meaning, in Wikipedia articles, potentially over-inflating the performance metrics of systems tested on this benchmark.

The most popular alternative to the MUSE dataset is that compiled in~\cite{hubness}, typically referred to as DINU.
This dataset was compiled automatically from Europarl~\cite{europarl}, a compilation of European parliament proceedings in 21 languages, typically packaged for evaluation systems into four primary languages: English, Finnish, German and Spanish~\cite{lost-evaluation}.
Training translation pairs are split by word frequency into five buckets: 1-5K, 5-20K, 20-50K, 50-100K, 100K-200K.
Within each bucket, 1,500 testing pairs are selected, as well as non-overlapping sets of training pairs in sizes of 1K, 5K, 10K and 20K.

\subsection{Knowledge Graph Entity Alignment Benchmarks}

Benchmarking experiments on knowledge graph entity alignment typically required two or more source knowledge graphs with a degree of known overlapping entities. 
One such way of generating these dataset is to use knowledge graphs describing the same set of triples in multiple languages. 
The WK31 datasets are an example of this paradigm, containing knowledge graphs focusing on the DBPedia person domain across English, French and German.
The WK31 dataset comes in two widely utilized variants based on the number of aligned nodes, namely WK31-15k and WK31-120k. 
These datasets are additionally evaluated based on the number of \textit{inter-lingual links} (ILL), where the ILL identify the same entity across two pairs of languages. 
The ILLs are typically used for the testing set for many evaluations and account for a small amount of the overall dataset, representing the challenge of generating a trustworthy seed dataset for supervised methods and motivating the search for many semi- or unsupervised solutions. 
Experiments on this dataset are reported in~\cite{mtranse, ntam, sea}.
As noted in~\cite{kg-align-survey}, there is a great deal of variance in the number of triples, entities, relations and seeds described in several publications using the WK31 dataset.
Here, we provide a summary of this dataset in Table~\ref{Tab:WK31Data}, based on the metrics reported in~\cite{ntam}.

In addition to cross-lingual knowledge graph datasets, several studies have split larger graphs into smaller components, each of which has linking entities that may be used as seeds to re-unify the graph.
The advantage of these datasets is that the target graph is entirely known and well defined, allowing for the number of seed entities to be scaled up and down and thus helping to experimentally validate the necessary amount of `overlap' needed to effectively perform the alignment task.
One such group of datasets named DFB-1, DFB-2 and DFB-3, constructed by~\cite{iptranse}, are constructed by randomly sampling triples from Freebase while specifying an overlapping threshold (OT).
A summary of the DFB datasets is provided in~\ref{Tab:DFBData}.

The approach of splitting a larger graph into many subgraphs is also applied to cross-lingual knowledge graphs as well, as is the case with the DBP15k dataset~\cite{jape}. 
This dataset uses DBPedia entities in Chinese, English, French and Japanese, creating four sets of ILLs, each containing 15,000 seed pairs, and is experimented with in~\cite{jape, ake, gcn-ea, mmea, bootea}.
A detailed description of the dataset is provided in~\ref{Tab:DBPData}.

\begin{center}
\begin{table}
\captionof{table}{Seed Alignments of WK31 Datasets \label{Tab:WK31Data}}
\centering
\begin{tabular}{| c | c | c |}
 \hline
 Dataset & Aligned Entities & Aligned Relations \\
 \hline
  WK31-15k-En-De & 2,070 & 445 \\
  WK31-15k-En-Fr & 3,116 & 598 \\
  WK31-120k-En-De & 9,680 & 772 \\
  WK31-120k-En-Fr & 42,378 & 1,127 \\
 \hline
\end{tabular}
\end{table}
\end{center}

\begin{center}
\begin{table}
\captionof{table}{Overlapping Alignments of DFB Datasets \label{Tab:DFBData}}
\centering
\begin{tabular}{| c | c | c | c | c |}
 \hline
 Dataset & Relations & Entities & OT & Seeds\\
 \hline
  DFB-1 & 1,345 & 14,951 & 0.5 & 5,000 \\
  DFB-1 & 1,345 & 14,951 & 0.5 & 500 \\
  DFB-1 & 1,345 & 14,951 & 0.1 & 500 \\
 \hline
\end{tabular}
\end{table}
\end{center}

\begin{center}
\begin{table}
\captionof{table}{Metrics of DFP Datasets \label{Tab:DBPData}}
\centering
\begin{tabular}{| c | c | c | c | c | c |}
 \hline
 Dataset & Language & Entities & Relations & Triples & Seeds\\
 \hline
  DBP15k-ZH-EN & Chinese & 66,469 & 2,830 & 153,929 & 15,000 \\
  DBP15k-ZH-EN & English & 98,125 & 2,317 & 237,674 & 15,000 \\
  DBP15k-FR-EN & French & 66,858 & 1,379 & 192,191 & 15,000 \\
  DBP15k-FR-EN & English & 105,889 & 2,209 & 278,590 & 15,000 \\
  DBP15k-JA-EN & Japanese & 65,744 & 2,043 & 164,373 & 15,000 \\
  DBP15k-JA-EN & English & 95,680 & 2,096 & 233,319 & 15,000 \\
 \hline
\end{tabular}
\end{table}
\end{center}

\section{Summary and Open Questions}
\label{sec:conclusion}

To summarize, we conducted a survey of the literature on the task of aligning diverse embedding spaces output by various neural networks for creating low-dimensional representations of data.
These methods have been thoroughly studied in the spaces of both natural language and graphs.
The majority of these methods aim to learn alignment models between spaces of the same underlying data type, i.e. words-to-words or graphs-to-graphs, typically with the alignment meant to bridge the gap between languages.
We find that there is significantly less research in bridging the gap between unlike embedding spaces, for example sentence embeddings and knowledge graphs, which we believe will provide significant gains in the fields of information extraction and data integration.
By identifying this gap and outlining existing methodologies we hope this survey provides an entry point for other researchers with a shared goal of aligning embedding spaces of diverse data types.

\clearpage
\bibliographystyle{unsrt}
\bibliography{alignment}

\end{document}